\documentclass[a4paper,twoside]{article}

\usepackage{epsfig}
\usepackage{subcaption}
\usepackage{calc}
\usepackage{amssymb}
\usepackage{amstext}
\usepackage{amsmath}
\usepackage{amsthm}
\usepackage{multicol}
\usepackage{pslatex}
\usepackage{apalike}
\usepackage[algo2e]{algorithm2e}
\usepackage[bottom]{footmisc}
\usepackage{romannum}
\usepackage{algorithm}
\usepackage{algorithmic}
\usepackage{multirow}
\pdfoutput=1

\usepackage{SCITEPRESS}     % Please add other packages that you may need BEFORE the SCITEPRESS.sty package.

\usepackage[a4paper,margin=1in]{geometry}
\usepackage{fancyhdr}
\begin{document}

\thispagestyle{fancy} %add this
\title{End-to-end Steering for Autonomous Vehicles via Conditional Imitation Co-learning}

\author{\authorname{Mahmoud M. Kishky\sup{1}, Hesham M. Eraqi\sup{2}\thanks{The work was conducted prior to Hesham Eraqi joining Amazon.} and Khaled M. F. Elsayed\sup{1}}
\affiliation{\sup{1}Faculty of Engineering, Cairo University, Egypt.}
\affiliation{\sup{2}Amazon, Last Mile, USA}
%\email{\{first\_author, second\_author\}@ips.xyz.edu, third\_author@dc.mu.edu}
}


\abstract{Autonomous driving involves complex tasks such as data fusion, object and lane detection, behavior prediction, and path planning. As opposed to the modular approach which dedicates individual subsystems to tackle each of those tasks, the end-to-end approach treats the problem as a single learnable task using deep neural networks, reducing system complexity and minimizing dependency on heuristics. Conditional imitation learning (CIL) trains the end-to-end model to mimic a human expert considering the navigational commands guiding the vehicle to reach its destination, CIL adopts specialist network branches dedicated to learn the driving task for each navigational command. Nevertheless, the CIL model lacked generalization when deployed to unseen environments. This work introduces the conditional imitation co-learning (CIC) approach to address this issue by enabling the model to learn the relationships between CIL specialist branches via a co-learning matrix generated by gated hyperbolic tangent units (GTUs). Additionally, we propose posing the steering regression problem as classification, we use a classification-regression hybrid loss to bridge the gap between regression and classification, we also propose using co-existence probability to consider the spatial tendency between the steering classes. Our model is demonstrated to improve autonomous driving success rate in unseen environment by 62\% on average compared to the CIL method.}

%Following the CIL work, we explored different methods to improve End-to-End steering for autonomous vehicles, In this paper, we propose a modified CIL network architecture to overcome the lack generalization issue, we also propose posing the steering regression problem as classification adopting two approaches, the first approach uses a combination of cross-entropy and mean squared error losses, the second approach involves considering the spatial relationship between the steering classes using the co-existence probability matrix to force a desired predefined distribution at the output layer, we used CARLA simulator for data collection and model evaluation. Our model showed autonomous  driving success rate improvement in unseen environment by 37\%. The source code will be shared publicly on github upon paper acceptance.}

% In this paper, we extend the CIL model to overcome its lack generalization by 
%- co-existence probability
%- co-learning
%- inverse encoding

%we also propose posing the steering regression problem as classification adopting two approaches, the first approach uses a combination of cross-entropy and mean squared error losses, the second approach involves considering the spatial relationship between the steering classes using the co-existence probability matrix to force a desired predefined distribution at the output layer, we used CARLA simulator for data collection and model evaluation. Our model showed autonomous  driving success rate improvement in unseen environment by 37\%. The source code will be shared publicly on github upon paper acceptance.

\onecolumn \maketitle \normalsize \setcounter{footnote}{0} \vfill

\section{\uppercase{Introduction}}
\label{sec:introduction}

An autonomous system is capable of understanding the surrounding environment and operating independently without any human intervention \cite{Autonomous_Robots}, in the context of autonomous vehicles, the objective is to mimic the behaviour of a human driver. To mimic a human driver's behaviour, the system is expected to take actions similar to those taken by a human driver in the same situation, the driver's action can be defined as a set of vehicle's controls such as steering, throttle, brake and gear.
\par
Autonomous driving systems can follow either the modular approach, or the end-to-end approach\cite{survey1}. In the modular approach, the system's pipeline is split into several components, each component has its own subtask, then the information provided by each component is combined to help the system understand the surrounding environment \cite{master} so the system can eventually generate different actions. On the other hand, the end-to-end approach replaces the entire task of autonomous driving with a neural network, where the network is fed observations (the inputs from the different sensors) and produces the predicted actions (steering, throttle, brake, gear), the objective is to train the network to learn the mapping between the observations and the actions.
\par
The end-to-end approach was first introduced in \cite{nvidia}, a convolutional neural network (CNN) was trained to map the raw pixels from a front-facing camera directly to steering commands. Later, the end-to-end approach was adopted widely in research such as in \cite{newp1}, \cite{newp2}, \cite{cil}, \cite{urban},\cite{cirl} and \cite{dcil} due to the simplicity of the process of development and deployment. Also, the model is free learn any implicit sources of information and the researcher is only concerned with developing a network that receives the raw data and delivers the final output \cite{master}, unlike the modular approach, there are no human-defined information bottlenecks \cite{survey2}.

\par
%One problem with the end-to-end approach is that there is not always a single right action for the same observation, for the same observation, there might exist several right actions based on the driver's intention such as the direction intended at intersections.

\cite{cil} proposed using a branched network architecture as shown in Figure 1, where the network is fed the navigational commands (go left, go right, go straight, follow lane) from a route planner representing the driver's intention, each specialist branch is dedicated to learn the mapping between the observations and the vehicle's actions independently. At test time, the navigational command acts as a switch to select the final action taken by the network. The major issue with the proposed model by \cite{cil} was its lack of generalization and the poor performance when deployed to unseen environment.
\par
In this work, we propose two contributions to improve the CIL end-to-end steering. In the first contribution, we introduce the conditional imitation co-learning (CIC) approach which involves modifying the CIL network architecture in \cite{cil}. \cite{cil} assumed total independence between the specialist branches while training, each branch was only trained on a subset of the training scenarios, for instance, the specialist branch dedicated to learn the right turns was only exposed to right turns scenarios during training. If the training data was not big enough to cover all the scenarios for all the branches, it could lead to unbalanced learning, which means that the model may perform properly in right turns and perform poorly in left turns or vice verse \cite{master}. We claim that one branch can make use of the features extracted by another. So, a branch dedicated to learn right turns can learn from observations collected in left turns and vice verse which enhances the model's generalization and increases its robustness in unseen environments.

\par
The second contribution is posing regression problem as classification, posing regression problem as deep classification problem was introduced in \cite{dex}, classification showed improvement in age prediction from a single image compared to regression. In our work, we adopt posing regression problem as classification as introduced in \cite{dex}, the classes were obtained by steering discretization. However, using this approach assume full independence between the steering classes ignoring their spatial tendency. So, we propose two improvements to the classification approach, the first improvement uses a combination of the categorical cross-entropy and the mean squared error losses to bridge the gap between classification and regression.
\par
The second improvement proposes considering the spatial relationship between the steering classes at the output layer using co-existence probability matrix, we claim that considering the spatial relationship between the classes will help to improve the overall performance of the model since the network will tend to predict the spatially close classes together.

\section{\uppercase{Related work}}

\subsection{Conditional Imitation Learning}
End-to-end Imitation learning aims to train the model to mimic an expert, the model with parameters \(w\) is fed a set of observations and actions \((o, a)\) pairs obtained from the expert. The model is optimized to learn the mapping function between the observations and actions \(F(o, w)\). In the context of autonomous driving, the observations are the data collected from different sensors (Cameras, Radars, LiDARs, ..), the actions are the vehicles controls such as steering, throttle and brake. The model is trained to mimic the actions taken by the expert to perform the task of autonomous driving.

The problem with the imitation learning approach is that the same observation could lead to different actions, based on the intention of the expert. Hence, the model cannot be trained to find a mapping function between the observations and the actions since it contradicts with the mathematical definition of the function itself, a function \(f:O \rightarrow A\) shall only map observation instance \(o\) to a single action \(a\). In other words, the driver's intention must be taken into consideration to make it mathematically possible to have a mapping function between the observations and the actions. So, at time step \(i\), the predicted action becomes a function of the driver's intention \(h_{i}\) as well as the observation \(a_{i} = F(o_{i}, h_{i})\), the network uses the navigational command \(c_i\) at intersections coming from a route planner to represent the driver's intention.

\cite{cil} proposed conditional imitation learning (CIL) approach by adopting a branched architecture as shown in Figure 1. After the feature extraction phase, the model is split into specialist branches each branch is dedicated to learn the mapping function between the observations and the actions given the navigational command (go left, go right, go straight, follow lane). Hence, the network produces the predicted action \(F\) at time step \(t\) given navigational command \(c_t\):
\begin{equation*}
F(o_{t}, c_{t}) = A^{c_{t}}(o_{t})
\end{equation*} where \(A^{c_{t}}\) is the predicted action by the specialist branch dedicated for command \(c_{t}\), which means that at testing, the navigational command is used as a switch to select the proper action to be taken by the model.

\begin {figure}
    \centering
    \includegraphics[width=1.1 \linewidth, height= 0.8 \linewidth]{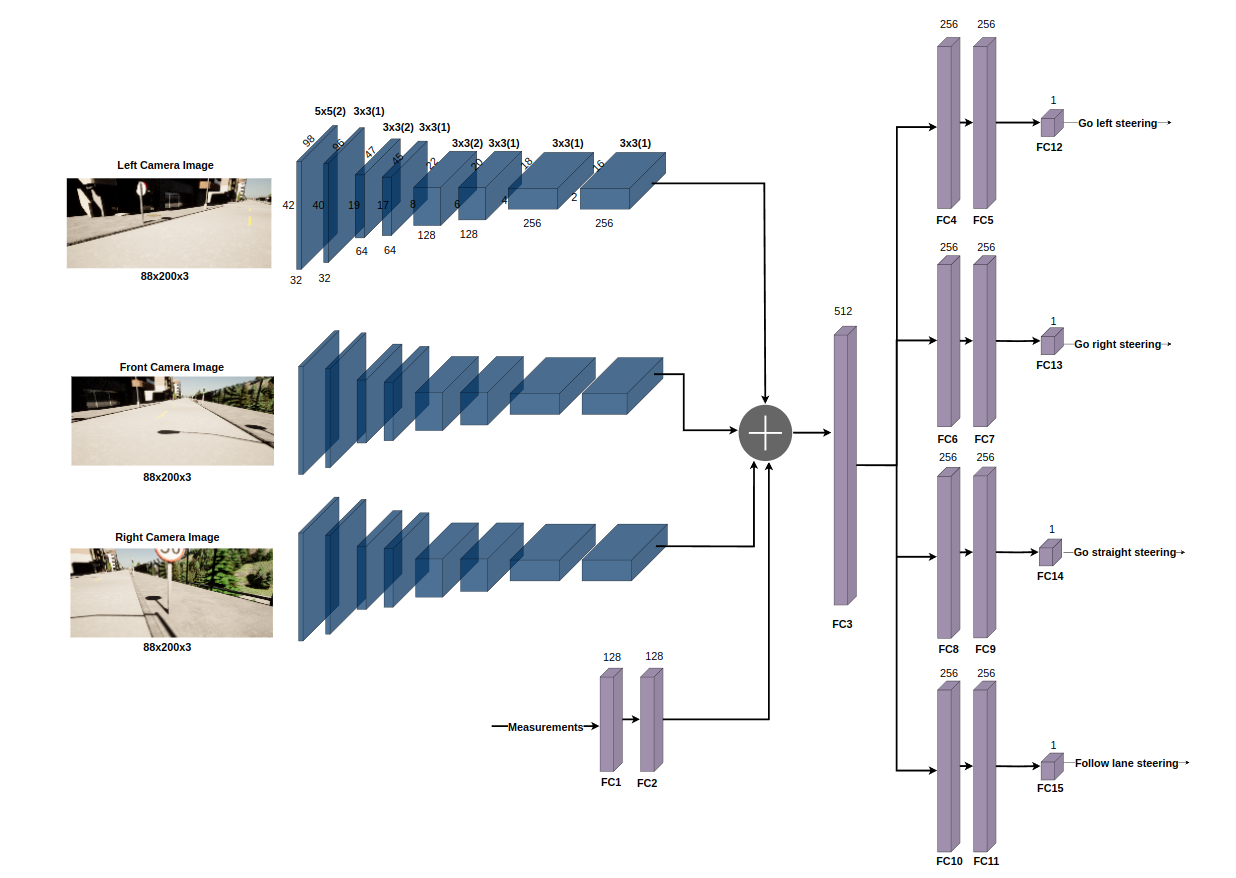}
    \caption{Network Architecture proposed by \cite{cil}}
    \label{fig:my_label3}
\end{figure}

\subsection{Regression as Classification}
In \cite{dex}, the regression problem of age prediction from images was posed as classification, the continuous age value was discretized to obtain the classification labels. Considering only the age values from 0 to 100, the network was trained to predict the true age of the human face in the input image. \cite{sine} followed the same approach to solve autonomous vehicle steering problem. Inspired by \cite{patent} and \cite{dex}, \cite{sine} proposed posing steering angle prediction for autonomous vehicles regression problem as classification, the model was trained and validated using comma.ai dataset \cite{coma}. \cite{sine} also proposed considering the spatial relationship between the steering angle classes using arbitrary function encoding, the encoding function was chosen to be a sine wave and the steering angle to be its phase shift. According to \cite{sine}, this choice of sinusoidal encoding led to gradual change in the values of the activations at the output layer. As shown in Figure 2, least squares error regression was used to optimize the model to generate a predicted waveform similar to the actual one from the steering angle encoding.

\begin {figure}
    \centering
    \includegraphics[width=0.8 \linewidth, height= 0.5 \linewidth]{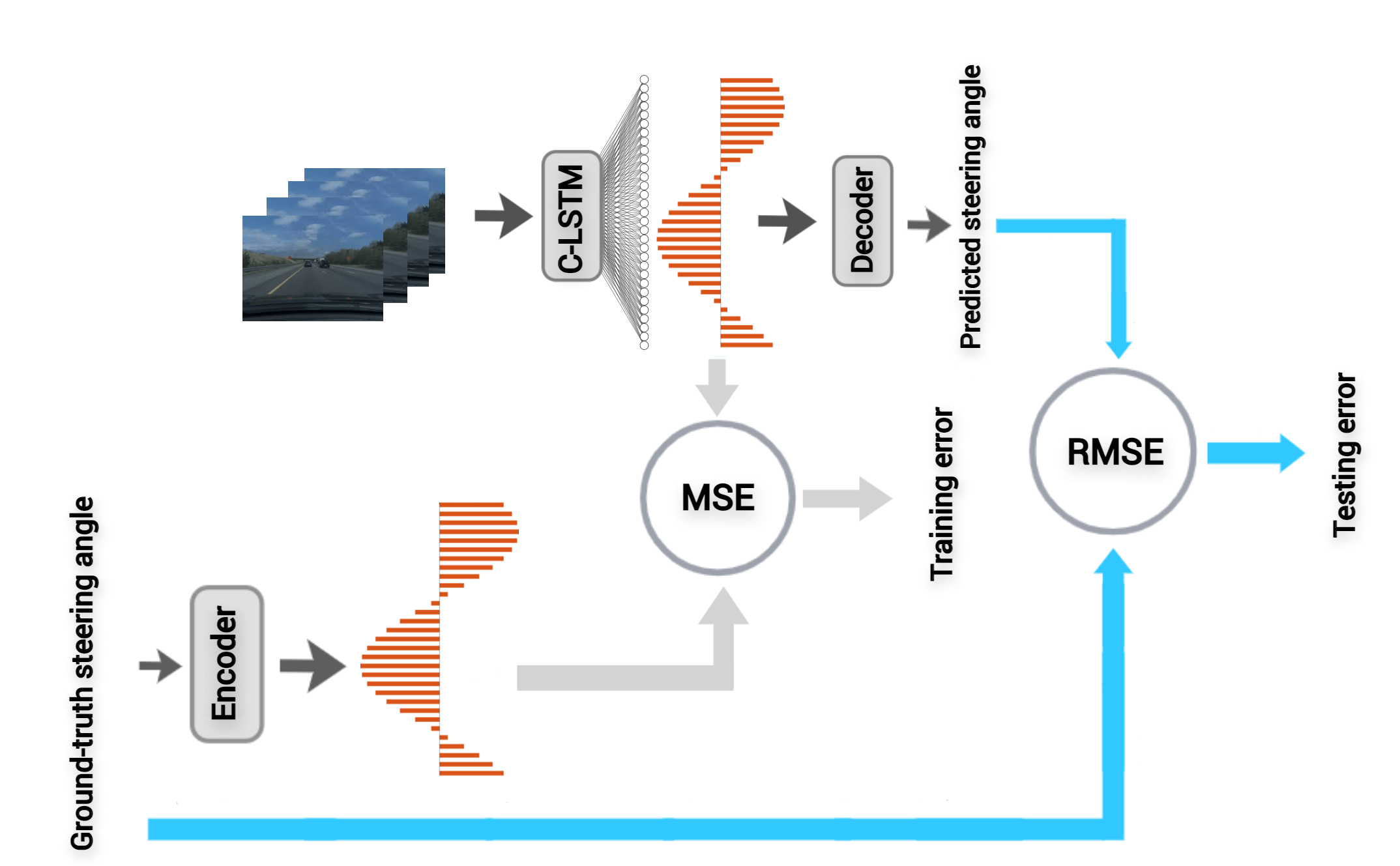}
    \caption{Training and deployment paths in the work in \cite{sine}}
    \label{fig:my_label2}
\end{figure}

The motive behind the need to consider the spatial relationship between the steering classes is the fact that the crossentropy loss only penalizes the model if the input is mislabeled neglecting all the scores corresponding to the other labels, which means that in case of mislabeling, the network will be penalized with the same amount no matter how big the difference between the predicted steering angle and the ground truth steering angle is. Since classification is only used to mimic regression, it is important to insure that the model still gives predictions close to the actual steering even in case of mislabeling.

\section{\uppercase{Methodology}}

\subsection{Data Collection}
The dataset was collected with the help of CARLA simulator \cite{carla}, CARLA provides a set of pre-built maps vary in size and complexity called towns as well as predefined weather conditions to facilitate the development of autonomous driving systems. In our work, we adopted the same three camera model proposed in \cite{cil}. Thus, three cameras were attached (front camera, right camera, left camera) to the ego-vehicle with resolution 200 x 88 pixels each. During data collection, we relied on CARLA simulator's autopilot feature, this feature allows the ego-vehicle to follow the lane and take random turns (left, right, straight) at intersections based on the navigational commands coming from the ego-vehicle's navigation agent, we ran 100 data collection episodes with the vehicle on autopilot mode with episode predefined period of ten minutes.

Besides the cameras' captures, we recorded the vehicle's steering and measurements (location and rotation) as well as the navigational commands from the ego-vehicle's navigation agent, the data was collected in Town01 with ClearNoon and ClearSunset weather conditions and sampled every 0.1 second, the camera captures were passed thought a randomized sequence of augmentation methods such as additive Gaussian noise, change of brightness and image cropping. As mentioned in \cite{cil}, noise was injected into the steering during data collection, the injected noise simulates the ego-vehicle's drifting and recovery to provide the network with examples of recovery from disturbances and allow the model to recover after making wrong predictions during testing.

Although simulation environments fail to capture the real world complexities, it provides a fast and safe way to develop and test autonomous driving models, Sim-to-Real is also a way to bridge the gap between simulation and real-world data by transferring the learned policies from simulated data into the physical world \cite{sim2real1}, \cite{sim2real2}.

\subsection{Specialist Branches Co-learning}

As discussed in Sec.\Romannum{1}, the CIL model introduced in \cite{cil} assumed independence between the specialist branches, each branch was trained on a subset of the training data to learn the mapping between the observations and actions given the navigational command coming form the route planner, the problem with this assumption is that it might lead to unbalanced learning, the dataset might contain enough scenarios for some branches to fit a mapping function and does not for the others. In our work, we propose considering dependencies between the specialist branches which allows the branches to co-learn by sharing their extracted features with each other.

As shown in Figure 3, assuming we have an N x 1 vector \({\hat{A}}_{t}\) representing the extracted features from the specialist branches at any time stamp \(t\), where N in the number of specialist branches, given an N x N co-learning matrix \(C_{t}\), for action prediction, we propose the following formula:
\begin{equation}
A_{t} = C_{t}\hat{A}_{t}
\end{equation} For four breaches dedicated for the navigational commands (go left, go right, go straight, follow lane) to which we will refer as l, r, s, f respectively:

\begin{equation*}
C_{t} = \begin{bmatrix} 1 & {c^{lr}}_{t} & {c^{ls}}_{t} & {c^{lf}}_{t}\\
                        {c^{rl}}_{t} & 1 & {c^{rs}}_{t} & {c^{rf}}_{t}\\
                        {c^{sl}}_{t} & {c^{sr}}_{t} & 1 & {c^{sf}}_{t}\\
                        {c^{fl}}_{t} & {c^{fr}}_{t} & {c^{fs}}_{t} & 1\\
                        \end{bmatrix}
\end{equation*}
\begin{equation*}
A_{t} = \begin{bmatrix} 
{\hat{A^{l}}}_{t} + {c^{lr}}_{t}\hat{{A^{r}}}_{t} +{c^{ls}}_{t}\hat{{A^{s}}}_{t} + {c^{lf}}_{t}\hat{{A^{f}}}_{t} \\

{\hat{A^{r}}}_{t} + {c^{rl}}_{t}\hat{{A^{l}}}_{t} + {c^{rs}}_{t}\hat{{A^{s}}}_{t} + {c^{rf}}_{t}\hat{{A^{f}}}_{t} \\

{\hat{A^{s}}}_{t} + {c^{sl}}_{t}\hat{{A^{l}}}_{t} + {c^{sr}}_{t}\hat{{A^{r}}}_{t} + {c^{sf}}_{t}\hat{{A^{f}}}_{t}\\

{\hat{A^{f}}}_{t} + {c^{fl}}_{t}\hat{{A^{l}}}_{t} + {c^{fr}}_{t}\hat{{A^{r}}}_{t} + {c^{fs}}_{t}\hat{{A^{s}}}_{t} \\

\end{bmatrix}
\end{equation*}, each matrix coefficient \(c^{ij}\) represents the the relationship between the extracted features \(\hat{A}^{i}\) and \(\hat{A}^{j}\) coming from branches \(i\) and \(j\) respectively.

%Each row in the co-learning matrix \(C^{i}\) represents the relationship between the the feature \(A^{i}\) coming form branch \(i\) and the other features coming from the other specialist branches.

In our work, we explored two approaches to generate co-learning coefficients. In the first approach, we break down the co-learning matrix as follows:
\begin{equation}
C_{t} = R \cdot E_{t}
\end{equation} where \(R\) is a hyperparameter binary matrix indicating the presence or absence of relationships between the specialist branches, the relationship matrix \(R\) is fine tuned and set manually before training. To produce matrix \(E\), we added a dedicated neural network to generate the co-learning coefficients corresponding to each branch, we chose \(tanh\) activation at the output layer to allow the co-learning matrix coefficients to vary between -1 and 1, the final predictions are then produced using Equation 1 as shown in Figure 3.

The second approach involves generating the co-learning coefficients directly from gated tanh units (GTUs) \cite{gtu} allowing the network to implicitly learn the relationship matrix \(R\). Gated units provide more control to the generated co-learning coefficients, the network is able to dynamically turn on/off the connections between the branches based on the driving scenario. The idea of the CIC approach can be extended to other applications especially those relying on foundational models, allowing the modules performing related tasks to share useful information.

\begin {figure}[H]
    \centering
    \includegraphics[width=0.85 \linewidth, height= 0.55 \linewidth]{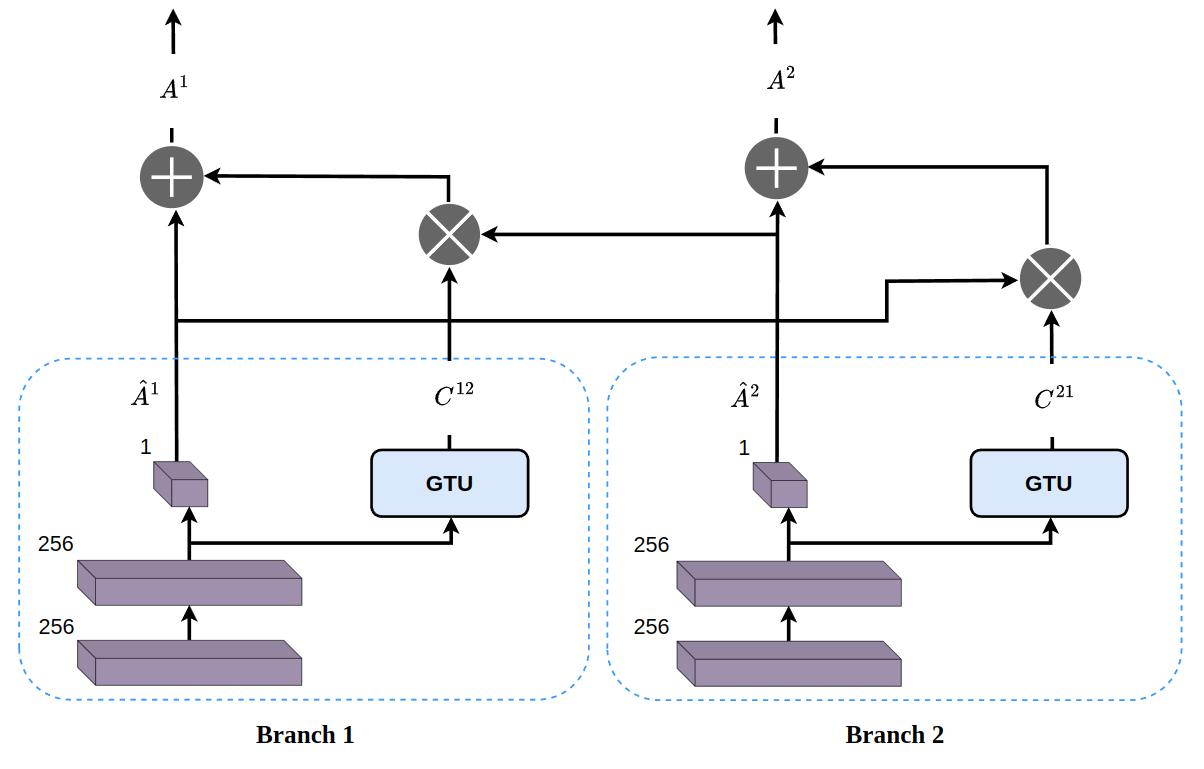}
    \caption{Co-learning between branch 1 and branch 2}
    \label{fig:co_learning}
\end{figure}

Other approaches addressed the improvement of multi-task learning (MTL) performance, such as mixture of experts (MoEs) \cite{moe}, soft parameter sharing, cross-stitch networks \cite{stitch} and sluice networks \cite{sluice}. MoEs involves adding a gated mixture of experts layer before splitting the network into specialist branches, allowing the branches to dynamically learn from a combination of experts instead of single shared network, each expert learns its own features from the input, then a gating network is used to decide which experts will be activated.

On the other hand, soft parameter sharing allows each task to have its own branch with its own parameters, then the model is regularized by L2 distance to encourage the parameters to be similar instead of learning from a shared network (hard parameter sharing). Unlike mixture of experts and soft parameter sharing approaches, \cite{stitch} and \cite{sluice} allow parameter sharing between specialist branches via learnable parameters.

\cite{stitch} introduced using cross-stitch units to learn the linear combinations of the activations coming from the different tasks at each layer of the network. In contrast, \cite{sluice} focused on determining which features should be shared between loosely coupled tasks, the hidden layers are split into two orthogonal subspaces, one for shared features and the other for task-specific features, the network learns to dynamically decide which features to be shared via learnable parameters.

%proposed learning linear combinations of the activation maps at each layer of the network via cross-stitch units, \cite{sluice} focused on determining which features should be shared between loosely coupled tasks, the hidden layers of specialist branches were split into two orthogonal subspaces, one for shared features and one for task-specific features, the network learn to dynamically decide which features to be shared via learnable parameters.

%focused more on learning the what features to be shared between loosely coupled tasks, the specialist branches' hidden layers are split into two orthogonal subspaces giving the network the flexibility to decide which features to be shared across the branches.

As opposed to other MTL approaches, our CIC model is output-oriented, we focus on learning the relationships between the outputs. In addition, we study the presence or absence of these relationships using GTUs, rather than trying to learn the relationships between hidden layer parameters. Our approach is motivated by the nature of our driving problem, where the overlapping between the branches' subsets is minimal and occurs only in a few specific driving scenarios, which reduces the network ability to learn the mapping between the parameters in the hidden layers and makes it less effective.

\subsection{Classification-Regression Hybrid Loss}

\cite{pose} proposed splitting the process of object pose prediction into two stages, coarse and refinement. The coarse stage involves a rough prediction of the pose using classification followed by offset prediction using regression, the network was spilt into two modules after feature extraction, one module for pose classification and the other for offset estimation. Given that the network had two types of outputs (softmax scores and offsets), \cite{pose} optimized the network using a combination of cross-entropy loss for classification and Huber loss \cite{huber} for regression.

In our work, we adopt a similar approach to \cite{pose}, for fair comparison between classification and regression, we only convert the regression output layer to a softmax layer rather than having two types of outputs in \cite{pose}. To bridge the gap between classification and regression, we used a combination of categorical cross-entropy (CCE) and mean squared error (MSE) losses for model optimization, the CCE loss imposes high penalty to the model is case of mislabeling allowing the network to produce coarse estimations to the steering, then MSE loss tunes the output activations for more accurate predictions.

Given \(O\) Nx1 the softmax output scores, \(y\) the true continuous steering, \(\overline{y}\) Nx1 the one hot representation of the discretized steering, \(m\) Nx1 steering midpoints corresponding to each class, N is the number of steering classes and the hyperparameter \(W\).

%\begin{equation}
%l = -\sum_{i = 1}^{N} {\overline{y}}_{i}\,log(O_{i}) + W \sum_{i = 1}^{N}
%{(y_{i} - E(O_{i}))}^{2}
%\end{equation} where

%\begin{equation*}
%E(O_{i}) =  \sum_{j=0}^{K-1} {O_{i}}^{j} m^{j} \\
%\end{equation*}

\begin{equation}
l = -\sum_{i=1}^{N} \overline{y}_i \log(O_i) + W  {( y - \hat{y})}^{2}
\end{equation} where

\begin{equation*}
\hat{y} = E(O) =  \sum_{i=1}^{N} {O_{i} m_{i}} \\
\end{equation*}

\subsection{Co-existence Probability Based Loss}

Inspired by \cite{dex} and \cite{sine}, this work poses vehicle steering regression problem as classification considering the spatial relationship between the steering classes. As illustrated in Sec.\Romannum{1}, it's important to insure that the model gives desirable predictions even in the case of mislabeling. Unlike \cite{sine}, we used co-existence probability matrix instead of sine wave encoding to represent this relationship, using co-existence probability matrix was introduced in \cite{co} to improve multi-class image categorization, this work used the same concept to force the scores at the output layer to follow a desired distribution which represents the spatial relationship between the discrete steering classes.

The co-existence probability matrix was used in \cite{co} to represent the statistical tendency of visual object to co-exist in images. In our steering problem, the objective of forcing the model to learn the output distribution is to make sure that in the case of misprediction, the model is always able to make acceptable predictions allowing the vehicle to recover from disturbances.

%For example in the scene in Fig.4, the model initially detected the road, the sky and a tree. Due to the use of co-existence probability matrix, the model understood the context and saw the high tendency of street light existence as well as other trees which enhanced the object detection accuracy of the model \cite{context}.

%\begin{figure}[h]
%    \centering
%    \includegraphics[width=0.8 \linewidth, height= 0.3 \linewidth]%{figures/coo.png}
%    \caption{Categorization system with co-existence statistics \cite{context}}
 %   \label{fig:my_label}
%\end{figure}

\par
%The co-existence probability matrix \(\mu\) is an N x N matrix, where N is the number of classes, the element \(\mu_{ij}\) represents the co-existence probability between labels \(i\) and \(j\),

%each row \(\mu_{i}\) represents the desired conditional probability distribution of the softmax scores \(O\) at the output layer as shown in Figure 5, given steering class prediction \(\hat{y} = i\):
%\begin{equation*}
%    \mu_{i} = p(O \,|\, \hat{y} = i) = p(O \,|\, argmax(O) = i)
%\end{equation*}

The co-existence probability matrix \(\mu\) is an N x N matrix, where N is the number of classes, the element \(\mu_{ij}\) represents the co-existence probability between labels \(i\) and \(j\), given that the output scores \(O \in \mathbb{R}^{N}\) and \(\mu \in \mathbb{R}^{NXN}\), the model shall try to find the output vector \(O\) that minimizes the following cost function according to \cite{co}:

\begin{equation}
l = -(1 - W) \sum_{i = 1}^{N} y_{i}\,log(O_{i}) - W \,O^{T}{\mu}\,O
\end{equation} the first term is the crossentropy loss, the second term describes how much the output scores vector \(O\) follows the desired distribution defined by \(\mu\), and \(W\) is a hyperparameter between 0 and 1 describing how much we care about forcing the desired distribution. Thus, the value \(O_{j}\) shall be pulled up or pulled down by \(O_{i}\) based on how high or low the value \(\mu_{i,j}\) is \cite{co}, the advantage of this approach is the flexibility it provides to the researcher to define the distribution most fitted for each class. In this work, we use Gaussian distribution with \({\sigma}^{2} = 1\).

\begin {figure}[h]
    \centering
    \includegraphics[width=0.8 \linewidth, height= 0.6 \linewidth]{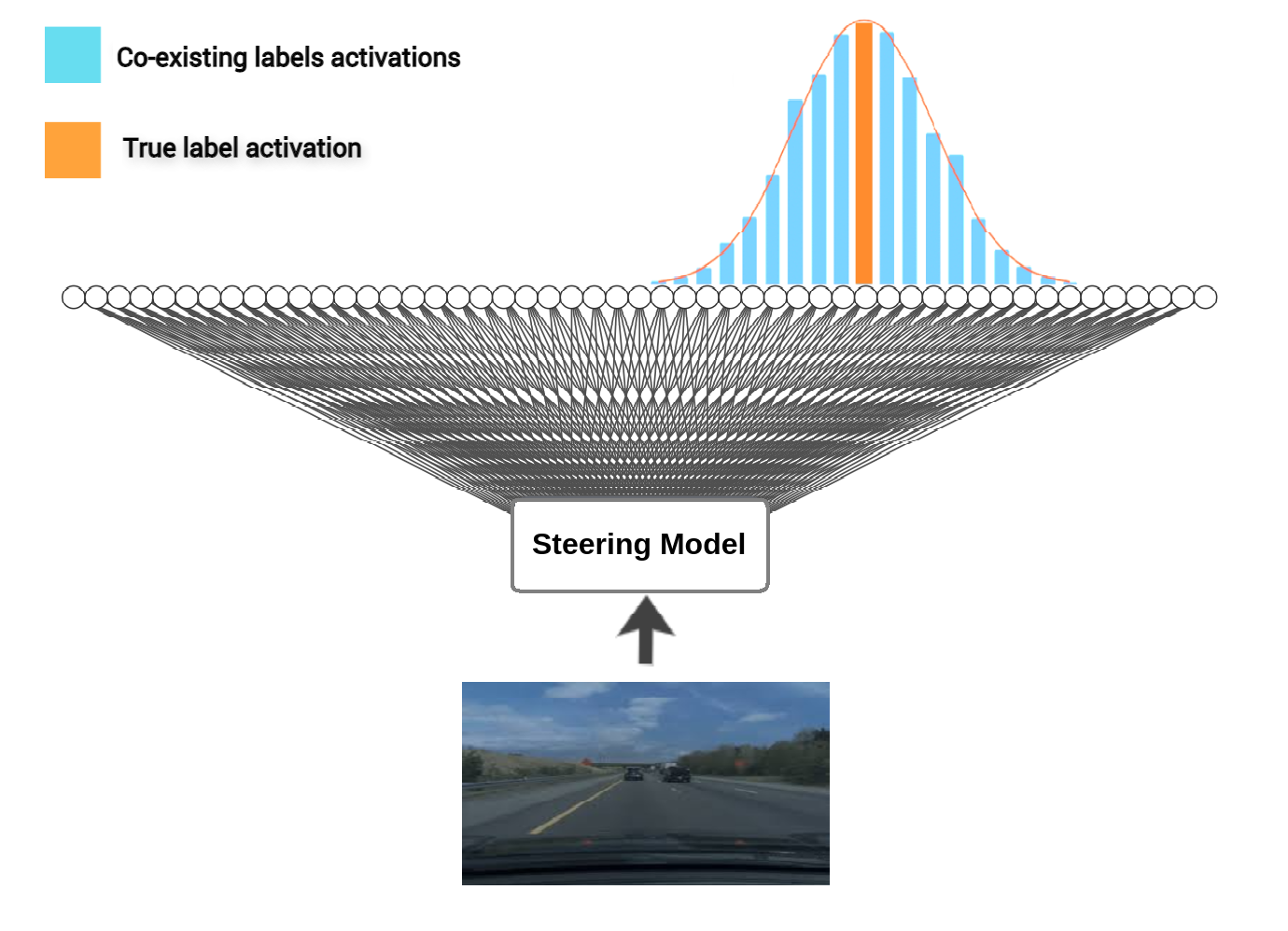}
    \caption{Spatial relationship between the steering classes}
    \label{fig:my_label4}
\end{figure}

\section{\uppercase{Training and Experiment}}
We split the dataset into training and validation sets, 70\% of the dataset for training and the remaining 30\% for validation, the models were trained using Adam optimizer \cite{adam} with \({\beta}_1 = 0.70\), \({\beta}_2 = 0.85\) and learning rate of 0.0002, we used mini-batches of 120 samples each, the mini-batches contained equal number of samples corresponding to each navigational command, the models were trained on the dataset collected as illustrated in Sec.\Romannum{3}. The steering values coming from the dataset samples ranges between -1 and 1, where -1 means a full turn to left and 1 means a full turn to the right, the actual value of steering depends on the vehicle used \cite{carla}. While training, the samples with steering between -0.8 and 0.8 were only considered. For classification models, we discretized the steering to 9 classes with 0.2 discreteization step.

The models were tested using CARLA simulator, we adopted a modified CoRL 2017 benchmark \cite{carla} where we considered only single turn task for performance evaluation, the task of the ego-vehicle is to successfully navigate from a start point and reach a destination point given predefined navigational commands forcing the vehicle to perform one single turn on its way. The vehicle was tested in in Town01 (training town) and Town02 (new town), in four different weather conditions, two training conditions (ClearNoon and ClearSunset) and two new conditions (midRainyNoon and wetCloudySunset), the new town and weather conditions are fully unseen to the steering models during training. We defined 38 pairs of start and destination points in Town01 and 40 pairs in Town02 associated with predefined navigational commands to cover all the intersections in both towns, which gives us 312 testing scenarios. On testing, we got the shown results in Tables \Romannum{1}, \Romannum{2}, \Romannum{3} and \Romannum{4}.

\setlength{\arrayrulewidth}{0.6mm}
\renewcommand{\arraystretch}{1.5} % Adjust row height

\begin{table}[H]
\centering
\caption{Driving Reach Destination Success Rate for Co-learning Model}
\resizebox{0.48 \textwidth }{!}{%
\begin{tabular}{|c|c|c|c|c|c|}
\hline
\multirow{2}{*}{Model} & \multicolumn{4}{c|}{Reach destination success rate}
\\ \cline{2-5}
                       & \multicolumn{2}{c|}{Training town} & \multicolumn{2}{c|}{New town} \\ \cline{2-5} 
                       & Training weathers & New weathers & Training weathers & New weathers \\ \hline
\hline
Regression (CIL)& 94.74 & 86.84 &  52.50 & 46.25\\
\hline
\textbf{GTU} & \textbf{100.00} & \textbf{93.42}  & \textbf{81.25} & \textbf{75.00} \\
\hline
\textbf{Fine Tuning} & \textbf{97.37} & \textbf{92.11} & \textbf{78.75}& \textbf{78.75}\\
\hline
\end{tabular}
}
\end{table}

\renewcommand{\arraystretch}{1.5} % Adjust row height
\begin{table}[H]
\centering
\caption{Driving Reach Destination Success Rate for Classification-Regression Hybrid Loss}
\resizebox{0.48 \textwidth}{!}{%
\begin{tabular}{|c|c|c|c|c|c|}
\hline
\multirow{2}{*}{Model} & \multicolumn{4}{c|}{Reach destination success rate}
\\ \cline{2-5}
                       & \multicolumn{2}{c|}{Training town} & \multicolumn{2}{c|}{New town} \\ \cline{2-5} 
                       & Training weathers & New weathers & Training weathers & New weathers \\ \hline
\hline
Classification& 72.37 & 43.42 &  41.25 & 32.50\\
\hline
W = 5 & 71.05 & 67.11  & 42.50 & 41.25 \\
\hline
\textbf{W = 10}& \textbf{97.37} & \textbf{85.53} & \textbf{64.25} & \textbf{56.25}\\
\hline
W = 15 & 86.84 & 50.52 & 50.00  & 41.25 \\
\hline
\end{tabular}
}
\end{table}

\begin{table}[H]
\centering
\caption{Driving Reach Destination Success Rate for Classification with Co-existence Probability Based Loss Model}
\resizebox{0.48 \textwidth}{!}{%
\begin{tabular}{|c|c|c|c|c|c|}
\hline
\multirow{2}{*}{Model} & \multicolumn{4}{c|}{Reach destination success rate}
\\ \cline{2-5}
                       & \multicolumn{2}{c|}{Training town} & \multicolumn{2}{c|}{New town} \\ \cline{2-5} 
                       & Training weathers & New weathers & Training weathers & New weathers \\ \hline
\hline
Classification& 72.37 & 43.42 &  41.25 & 32.50\\
\hline
W = 0.4 & 71.05 & 65.79 & 45.00 & 36.25\\
\hline
W = 0.6 & 76.32 & 68.42 & 63.75 & 48.75\\
\hline
W = 0.8 & 69.74 & 61.84 & 50.00 & 41.25\\
\hline
\end{tabular}
}
\end{table}

%\renewcommand{\arraystretch}{1.5} % Adjust row height
%\begin{table}[H]
%\centering
%\caption{Driving Reach Destination Success Rate For Steering Sinewave Encoding Model}
%\resizebox{\textwidth / 2}{!}{%
%\begin{tabular}{|c|c|c|c|c|c|}
%\hline
%\multirow{2}{*}{Model} & \multicolumn{4}{c|}{Reach destination success rate}
%\\ \cline{2-5}
%                       & \multicolumn{2}{c|}{Training town} & \multicolumn{2}{c|}{New town} \\ \cline{2-5} 
%                       & Training weathers & New weathers & Training weathers & New weathers \\ \hline
%\hline                       
%N = 21 &  &   &  & \\
%\hline
%N = 61 &  &  & & \\
%\hline
%N = 101& &  & & \\
%\hline
%N = 141&  &  & &\\
%\hline
%\end{tabular}
%}
%\end{table}

\renewcommand{\arraystretch}{1.5} % Adjust row height
\begin{table}[H]
\centering
\caption{Driving Reach Destination Success Rate (All Models)}
\resizebox{0.48 \textwidth}{!}{%
\begin{tabular}{|c|c|c|c|c|c|}
\hline
\multirow{2}{*}{Model} & \multicolumn{4}{c|}{Reach destination success rate}
\\ \cline{2-5}
                       & \multicolumn{2}{c|}{Training town} & \multicolumn{2}{c|}{New town} \\ \cline{2-5} 
                       & Training weathers & New weathers & Training weathers & New weathers \\ \hline
\hline                      
\textbf{Regression (CIL)}& \textbf{94.74}& \textbf{86.84}& \textbf{52.50} &\textbf{46.25}\\
\hline
\textbf{Co-learning (GTU)} & \textbf{100.00} & \textbf{93.42}  & \textbf{81.25} & \textbf{75.00} \\
\hline
\textbf{Parameter soft sharing} & \textbf{96.05} & \textbf{84.21} & \textbf{67.50} & \textbf{60.00 }\\
\hline
\textbf{Mixutre of experts} & \textbf{97.37} & \textbf{81.58} & \textbf{72.50} & \textbf{67.50}\\
\hline
\textbf{Sluice network} & \textbf{100.00} & \textbf{89.47} & \textbf{70.00} & \textbf{61.25}\\
\hline
Classification& 72.37 & 43.42 &  41.25 & 32.50\\
\hline
Co-existence based loss &76.32 & 68.42 & 63.75 & 48.75\\ 
\hline
\textbf{CCE + MSE} & \textbf{97.37} & \textbf{85.53} & \textbf{64.25} & \textbf{56.25}\\
\hline
%\textbf{Arcsine encoding} & \textbf{97.37} & \textbf{90.79} & \textbf{80.00} & \textbf{72.50}\\
%\hline
%CCE + MSE + GTU & &  & & \\
%\hline
\textbf{Sine wave encoding} & \textbf{97.37}& \textbf{86.84}& \textbf{57.50}& \textbf{48.75}\\
\hline
%Arcsine encoding + co-learning & &  & & \\
%\hline
\end{tabular}
}
\end{table}

%\begin{figure}[H]
%    \centering
%    \includegraphics[width= 0.95 \linewidth, height= 0.7\linewidth]{figures/carla_map.png}
%    \caption{High level commands from CARLA, follow lane (blue), go right (green) and go left (yellow),
%    the height level commands determine the path followed by the ego-vehicle to reach the destination points (red dots)}
%    \label{fig:my_label}
%\end{figure}

As discussed in Sec.\Romannum{2}, the CIL model \cite{cil} lacked generalization when tested in unseen environments. On the other hand, our proposed CIC model tackled this issue, we could see improvement in the reach destination success rate in unseen environment (unseen town, unseen weather) by 62\%. Classification model failed to improve the CIL performance which conforms to the results in \cite{dex} where classification could only outperform regression for some datasets. Using the hybrid loss in Equation 3 showed 21\% improvement compared to the CIL model. Nevertheless, the co-existence based loss in Equation 4 only showed improvement compared to the basic classification model but failed to outperform the CIL regression model. MTL approaches such as MoEs \cite{moe}, soft parameter sharing and sluice networks \cite{sluice} showed improvement to the CIL model especially in new town, but failed to outperform our CIC model while stitch network \cite{stitch} failed to learn the driving task, sine wave encoding \cite{sine} also showed slight improvement compared to the CIL model.

%we can notice that once we had enough neurons at the output layer to draw a smooth sine wave, the model's performance was almost identical.

%On the other hand, classification with co-existence based samples relabeling also showed 22\% improvement unseen environment compared to the CIL model. Nevertheless, the co-existence bases loss in Eq.4 only showed improvement compared to the basic classification model but failed to outperform the CIL regression model. Sine wave encoding did not show any improvement compared to the CIL model's performance, we also noticed that once we had enough neurons at the output layer to draw a smooth sine wave, the model's performance was almost the same.

\section{\uppercase{Conclusion}}
In this work, we propose two contributions to the end-to-end steering problem tackled by the conditional imitation learning (CIL) model, the CIL model suffered from lack of generalization and poor performance when tested in unseen environment, the first contribution of this work is conditional imitation co-learning (CIC), the introduced approach proposes a modified network architecture that allows the specialist branches in the CIL model to co-learn to overcome the generalization issue and increase the model's robustness in unseen environment, the other contribution is posing the steering regression problem as classification by using a combination of CCE and MSE losses. The CIC model showed a significant improvement to performance in unseen environment by 62\% while posing regression as classification showed only improvement by 21\%.

%In the second method, we use a modified loss formula to consider the spatial relationship between the spatial relationship between the steering classes using co-existence probability matrix, this method only showed improvement compared to the basic classification model 
%considering the spatial relationship between the obtained steering classes using the co-existence probability matrix, the co-learning model improved the reach destination success rate in unseen environment by 37\%, the classification with labels reassignment improved it by 22\%. On the other hand, classification model with co-existence based loss only outperformed the basic classification model, the sine wave encoding model did not show any significant improvement over the CIL model.

\bibliographystyle{apalike}
{\small
\bibliography{refs}}

\end{document}